%% file: BLS.tex
\documentclass{article}                     
\usepackage[margin=3.3cm]{geometry}
\usepackage{amsmath}
\usepackage{amssymb}
\usepackage{graphicx}
\usepackage{bm}
\usepackage{natbib}
\usepackage[colorlinks,citecolor=black,linkcolor=black,urlcolor=black,filecolor=black]{hyperref}
 \usepackage{graphicx}
 \usepackage{import}
\usepackage{algorithm,algorithmic} 
\usepackage{booktabs} 
\usepackage{multirow} 
\usepackage{float}  
\usepackage{mathtools} 
\usepackage{xcolor}
\usepackage[title]{appendix}                                
 
\renewcommand\d[0]{\ensuremath{\operatorname{d}\!}}
\usepackage{authblk}

\author[1]{Ingvild M. Helg\o y \thanks{Corresponding author. E-mail address: Ingvild.Helgoy@uib.no}}
\author[1]{Yushu Li}
\affil[1]{University of Bergen, Bergen, Norway}
\date{\today}

\begin{document}
\title{A Bayesian Lasso based  Sparse  Learning Model
}
\maketitle

\begin{abstract}
The Bayesian Lasso is constructed in the linear regression framework and applies the Gibbs sampling to estimate the regression parameters. 
This paper develops a new sparse learning model, named the Bayesian Lasso Sparse (BLS) model, that takes the hierarchical model formulation of the Bayesian Lasso. The main difference from the original Bayesian Lasso lies in the estimation procedure; the BLS method uses a learning algorithm based on the type-II maximum likelihood procedure. Opposed to the Bayesian Lasso, the BLS provides sparse estimates of the regression parameters. 
The BLS method is also derived for nonlinear supervised learning problems by introducing kernel functions. 
We compare the BLS model to the well known Relevance Vector Machine, the Fast Laplace method, the Byesian Lasso, and the Lasso, on both simulated and real data. The numerical results show that the BLS is sparse and precise, especially when dealing with noisy and irregular dataset.

\end{abstract}


\section{Introduction}
\label{intro}

The Lasso of~\cite{tibshirani1996regression} is a state-of-the-art  method for solving linear regression problems.  It performs both estimation and variable selection since some of the estimated coefficients can be set to zero during the estimation procedure. 
It is known that the Lasso estimate  corresponds to a Bayesian posterior mode estimate where the priors for the  coefficients  are identical and independent  Laplace distributions~\citep{tibshirani1996regression, friedman2001elements}. 
Motivated by the connection between the Lasso and the Laplace prior in the Bayesian framework, \cite{park2008bayesian} introduced a fully Bayesian model of the Lasso. The Bayesian Lasso by  \cite{park2008bayesian} uses a Laplace prior conditioned on the variance of the random noise, while the approximation of the posterior thereafter is obtained by using the Gibbs sampler. 
\cite{park2008bayesian} argued that the prior conditioned on the variance is important because it guarantees a unimodal posterior. On the other hand, using the unconditional Laplace prior from the Lasso, the posterior can easily have more than one mode  which may slow down convergence of the Gibbs sampler and  less concise point estimates. 
A limitation of the Bayesian Lasso is that it is not a sparse model because none of the estimated  coefficient parameters  from the Gibbs sampling will be set exactly to zero. Hence, the Bayesian Lasso does not perform variable selection by default. However, \cite{park2008bayesian} suggest to use the corresponding credible intervals to guide variable selection, but this requires choosing appropriate threshold values. Other similar methods that attempt to overcome the problem of sparsity includes a Bayesian Elastic net method proposed by \cite{li2010bayesian}, and
the Bayesian adaptive Lasso by \cite{leng2014bayesian} that uses an adaptive penalty parameter which promotes sparsity. Both these methods utilize the Gibbs sampler to approximate the posterior distribution. \cite{hahn2015decoupling} review the relationship between variable selection priors and shrinkage priors  and present  an overview of several Bayesian model selection methods. 

%

The last decades, kernel-based machine learning methods have gained increased interest. One of the more popular kernel-based methods is the Support Vector Machine~(SVM)~\citep{boser1992training, vapnik1997support, scholkopf1999advances}, which is a sparse method that has proven to perform well for several different cases, see e.g., \cite{steinwart2008support}.
Despite its popularity, the SVM has also some known limitations.
The SVM is purely deterministic because the SVM prediction is a point estimate.  
Further,
\cite{tipping2001sparse} points out that the SVM model does require that the kernel must satisfy the Mercer conditions. To overcome this issue, among others, \cite{tipping2001sparse} presents a new method called the Relevance Vector Machine (RVM), and a faster version was later developed~\citep{tipping2003fast}. Opposed to the SVM method, the RVM is formulated in the Bayesian framework and the kernel does not have to satisfy the Mercer condition.
 Both SVM and RVM can be applied to solve nonlinear regression and classification problems. 
They can both achieve sparsity in the sample domain and depend only on a subset of the kernel functions and associated training samples. 
\cite{tipping2001sparse} demonstrates that the RVM is more sparse than the SVM, and
the RVM has shown great success in solving many machine learning and pattern recognition problems \citep[see, e.g.,][]{
ghosh2008statistical,liu2015lithium,kaltwang2015doubly,kong2019relevance, qiao2019novel, liu2020facies}. Several extensions of the RVM model can be found in the literature \cite[see, e.g.,][]{wipf2004sparse,krishnapuram2005sparse,schmolck2007smooth,ji2008bayesian, tien2018novel, agrawal2019ensemble}. The RVM has inspired a variety of similar Bayesian methods, including the Fast Laplace method (FLAP) by \cite{babacan2010bayesian}, which has mainly been applied in the field of compressive sensing for signal reconstruction. In the derivation of the FLAP method, \cite{babacan2010bayesian} show how the variance can be estimated in the optimization procedure. However, in the numerical results they use a fixed value for the variance because of unstable estimates when the method was used on compressive sensing datasets.

This paper presents an adaptation of the Bayesian Lasso by \cite{park2008bayesian} to the kernel-based framework  such that general nonlinear problems can be solved and sparsity in sample domain is achieved.
  We call this new model the Bayesian Lasso Sparse (BLS) model.
Instead of using the Gibbs sampler in \cite{park2008bayesian} to estimate the parameters, we adapt the  marginal likelihood maximization method (also known as type-II maximum likelihood) from the RVM~\citep{tipping2003fast}.
The BLS method assigns individual weight parameters to each input sample and the hierarchical structure from \cite{park2008bayesian} is utilized in the BLS model which results in an “automatic relevance determination"~(ARD)  conditional prior \citep{mackay1992evidence} for the weight
parameters.
We will show that this approach leads to a sparse solution in the sample domain.
 In addition, we provide  an analysis of how the variance of the random noise  affects the hyperparameter estimation by using the conditional prior for the weight parameters.
We conduct a comprehensive simulation study to compare the performance of the BLS  with the fast RVM method of \cite{tipping2003fast} and the FLAP of~\cite{babacan2010bayesian}. We also include the version of the FLAP model that estimate the variance to test its performance on non-linear regression data.

 

This paper also adjusts the BLS method to achieve sparsity in the variable domain in the original linear regression framework by \cite{park2008bayesian}. The difference between the  BLS in the linear regression framework  and the original Bayesian Lasso  by \cite{park2008bayesian} lies at the estimation procedure. The  type-II maximum likelihood estimation   results in a sparse model for the BLS where the most important variables  have been selected. The  Gibbs sampling in the Bayesian Lasso will in general only produce shrinkage of the coefficients so all the input variables remain in the model. Simulation studies are used to investigate the sparsity of the BLS in variable domain, and compare it to the Bayesian Lasso and the original Lasso.

The remainder of the paper is divided into the following sections: Section~\ref{sec:bls}  contains a detailed description of the BLS method, including a fast optimization algorithm and theoretical derivation of the variance related threshold for the hyperparameters.  Section~\ref{sec:results} presents results from simulation studies and compare the BLS with other kernel-based learning methods where sparsity is achieved in the sample domain. Section~\ref{sec:variablesel} shows how the  BLS method can be used for variable selection, and concluding remarks are given in Section~\ref{sec:conclusion}.

\section{The Bayesian Lasso Sparse Model } \label{sec:bls}
After a general description of supervised learning,
 this section explains in detail the hierarchical structure of the BLS method and its kernel framework. In addition, we specify the type-II maximum likelihood method, as well as an analysis of how the variance of the random noise can affect the sparsity. Inference and  prediction by using the  BLS model will be briefly described at the end of this section.

\subsection{Supervised learning} \label{sec:methodology}
Supervised learning contains a set of training data $\{\boldsymbol{x}_i,y_i \}_{i=1}^N$, where $\boldsymbol{x}_i \in \mathbb R^D$ is a $D$-dimensional input vector and $y_i \in \mathbb R$  is the corresponding scalar target value. Based on the training data, we aim at constructing a function $f(\boldsymbol{x})$ that can model the underlying relationship between the input covariate $\boldsymbol{x}_i$  and the target observation $y_i$.  A common way to construct  $f(\boldsymbol{x})$ is to use a set of $M$ linearly independent basis vectors, $\phi_m$:
\begin{equation}
\label{fxapro}
{f}(\boldsymbol{x}) = w_0+\sum_{m=1}^M w_m\phi_m(\boldsymbol{x}),
\end{equation}
where $\boldsymbol{w} = (w_0,w_1, \dots, w_M )^\top$ is a vector of weight parameters.

We assume that the observed targets, $y_i$, are samples of the function $f(\textbf x)$ with added noise which follows a Gaussian distribution. To relate the targets to the input, we first create a design matrix, $\boldsymbol \Phi$, from the basis vectors as
\begin{eqnarray*}
\boldsymbol{\Phi} = [\boldsymbol 1, \boldsymbol{\phi}_1, ..., \boldsymbol{\phi}_M],
 \qquad \boldsymbol{\phi}_m =  (  \phi_{m}(\boldsymbol{x}_1), \dots , \phi_{m}(\boldsymbol{x}_N) )^\top, \quad m=1,\ldots,M.
\end{eqnarray*} 
Let $\boldsymbol{\epsilon} = ( \epsilon_1, \dots, \epsilon_N )^\top$ be the standard Gaussian distributed noise vector, we then have:
\begin{equation} \label{linmod}
  \boldsymbol{y} = \boldsymbol{\Phi}\boldsymbol{ w } + \boldsymbol{\epsilon} ,   \qquad \boldsymbol{\epsilon}\sim \mathcal{N}( \boldsymbol 0, \sigma^2\boldsymbol{I}_N),
\end{equation}
where $\textbf y = (y_1, \ldots, y_N)^\top$, $\sigma^2$ is the variance of the error terms that are normally distributed and $\boldsymbol I_N$ is the $N\times N$ identity matrix.
\color{black}
%

\subsection{Model specification and parameter estimation}
The first step to construct the BLS model is to define the basis functions in Equation~\eqref{fxapro}.
 In the BLS method, the basis functions are defined by $M=N$ kernel functions $\phi_m(\boldsymbol{x}) =  K (\boldsymbol{x}, \boldsymbol{x}_m)$; $m=1, \dots, N$. The kernel function $K (\cdot,\cdot)$ is centred at each of the training input vectors. Thus, each basis function $\phi_m(\boldsymbol{x})$  corresponds to one training input vector $\boldsymbol{x}_m$. The Gaussian and the polynomial kernel functions are the most used kernel functions for sparse learning models.
%
%

 After the type of kernel function is chosen,  the learning task is  to estimate the weight parameters, $\boldsymbol{w}$,  from the training data.
Similar to the Bayesian Lasso, the conditional prior on the weight parameters is given by 
\begin{equation} \label{bpriorw2}
p(\boldsymbol w | \sigma^2, \lambda) = \prod_{i=0}^N \frac{\sqrt{\lambda}}{2\sqrt{\sigma^2}}e^{-\sqrt{\lambda} |w_i|/\sqrt{\sigma^2}}, \qquad  \lambda \geq 0,
\end{equation}
which can be   recognised as a Laplace prior conditioned on $\sigma^2$ and $\lambda$.  Further, the likelihood function
  is derived from Equation~\eqref{linmod} as
\begin{equation} \label{condy}
p(\boldsymbol{y} |\boldsymbol{w} ,\sigma^2 ) = \mathcal{N}(\boldsymbol{y} |\boldsymbol{\Phi}\boldsymbol{w}, \sigma^2).
\end{equation}

Bayesian inference is based on the posterior distribution of $\boldsymbol{w}$ given the data, $p(\boldsymbol{w|y},\sigma^2 , \lambda)$, which can be calculated from~Equations~\eqref{bpriorw2} and \eqref{condy}. 
However, the inclusion of the Laplace prior in Equation~\eqref{bpriorw2} makes an analytical solution intractable. We therefore proceed to use a hierarchical representation of the full model, similar to the Bayesian Lasso described by \cite{park2008bayesian}:
\begin{eqnarray}
 \label{blshiera}
\mathrm{Likelihood} \quad  p(\boldsymbol{y} |\boldsymbol{w} ,\sigma^2)  &=& \mathcal{N}(\boldsymbol{y} |\boldsymbol{\Phi}\boldsymbol{w}, \sigma^2), \\ 
\mathrm{Hierarchical \ prior} \quad p(\boldsymbol{w}| \boldsymbol\tau, \sigma^2)  &=& \prod_{i=0}^N \mathcal{N} (w_i | 0, \tau_i \sigma^2), \qquad \tau_i \geq 0, \label{blsw} \\  
\mathrm{Hyperprior} \quad p( \boldsymbol\tau | \lambda)  &=& \prod_{i=0}^N \frac{\lambda}{2} \mathrm{e}^{-\frac{\lambda \tau_i}{2}}, \hspace{1.68cm} \lambda \geq 0. \label{hyperpri}
\end{eqnarray}
The hierarchical prior in Equation~\eqref{blsw} can be viewed as an ARD  prior   where for $i=0,1,\dots , N$, there is an individual hyperparameter $\tau_i$ associated independently with each individual weight $w_i$.  Equation~\eqref{blsw} shows  that when a hyperparameter is estimated to be zero in this ARD prior, it will force the corresponding weight parameter to be zero, and prune the input vector and the related sample.

The priors for  $\lambda$ and $\sigma^2$ in Equations~\eqref{blshiera} -~\eqref{hyperpri}, are defined as the Gamma and  inverse Gamma distribution, respectively:
\begin{eqnarray} 
\label{priorlasso} p(\lambda) &=& \frac{b ^a}{\Gamma (a)} (\lambda)^{a-1} e^{-b \lambda}, \hspace{1.63cm} a,b \geq 0,\\ 
\label{priorsigma} p(\sigma^2) &=&  \frac{d^c}{\Gamma (c)} (\sigma^2)^{-c-1} e^{-d/ \sigma^2}, \hspace{1cm} c,d  \geq 0.
\end{eqnarray}
By combining Equations~\eqref{blshiera} -~\eqref{priorsigma} from the above hierarchical Bayesian model, we get the following joint distribution of the dataset, parameters and hyperparamters:
\begin{eqnarray*}
 p(\boldsymbol{y}, \boldsymbol{w}, \boldsymbol{\tau}, \sigma^{2}, \lambda)   =  p(\boldsymbol{y | w} ,\sigma^{2})p(\boldsymbol{w}|\boldsymbol{\tau}, \sigma^2)p(\boldsymbol{\tau}|\lambda) p(\lambda)p(\sigma^{2}).
\end{eqnarray*}
Given the observed data, the following posterior over all unknowns can be found:
\begin{equation}\label{bayes1}
p(\boldsymbol{w}, \boldsymbol\tau ,\sigma^2, \lambda | \boldsymbol{y}) = 
\frac { p(\boldsymbol{y}, \boldsymbol{w}, \boldsymbol{\tau}, \sigma^{2}, \lambda)} {p(\boldsymbol{y})}.
\end{equation}
When the posterior distribution in Equation~\eqref{bayes1} is available, an expression of the predictive distribution for the output $y^*$ can be obtained as
\begin{equation}\label{predy}
 p(y^*|\boldsymbol{y}) = \int p(y^* | \boldsymbol{w}, \sigma^2)  p(\boldsymbol{w}, \boldsymbol\tau, \sigma^2 , \lambda| \boldsymbol{y})   \d\boldsymbol{w}  \d\boldsymbol\tau  \d\sigma^2 \d\lambda.
\end{equation}
However, neither the posterior in Equation~\eqref{bayes1} nor the predictive  distribution in Equation~\eqref{predy} can be computed analytically as the normalising integral $ p(\boldsymbol{y}) =  \int p(\boldsymbol{y}, \boldsymbol{w}, \boldsymbol{\tau}, \sigma^{2}, \lambda) \d\boldsymbol{w}  \d\boldsymbol\tau  \d\sigma^2 \d\lambda$ is intractable. 
Instead, 
the BLS model applies the type-II maximum likelihood estimation from \cite{tipping2003fast} to approximate the predictive distribution in Equation~\eqref{predy}.

As the posterior in Equation~\eqref{bayes1} can not be found directly from Bayes' rule, we use the  decomposition:
\begin{equation} \label{decomp2}
p(\boldsymbol{w}, \boldsymbol\tau ,\sigma^2, \lambda | \boldsymbol{y})=p(\boldsymbol{w|y}, \boldsymbol\tau ,\sigma^2, \lambda)p( \boldsymbol\tau ,\sigma^2 ,\lambda | \boldsymbol{y}).
\end{equation}
The distribution $p(\boldsymbol{w|y},\boldsymbol\tau, \sigma^2, \lambda)$ on the right hand side of Equation~\eqref{decomp2}, is a Gaussian distribution with the following mean vector and covariance matrix:  \begin{eqnarray*}
\boldsymbol{\mu} &=& \sigma^{-2} \boldsymbol{\Sigma}\boldsymbol{\Phi}^T\boldsymbol{y}, \label{muy} \\
\boldsymbol{\Sigma} &=& [\sigma^{-2}\boldsymbol{\Phi}^T\boldsymbol{\Phi} + \Lambda^{-1}]^{-1}, \label{signy}
\end{eqnarray*}
where  $\Lambda = \mathrm{diag}(\tau_i\sigma^2)$.  

To estimate $\boldsymbol\tau$, we search for the local maximum of $p( \boldsymbol\tau ,\sigma^2 ,\lambda | \boldsymbol{y})$ in Equation~\eqref{decomp2} with respect to the  individual hyperparameters $\tau_i$ by using the type-II maximum likelihood procedure. 
To find the maximum, we use $p(\boldsymbol\tau ,\sigma^2 ,\lambda | \boldsymbol{y}) = p(\boldsymbol{y} ,\boldsymbol\tau ,\sigma^2 ,\lambda)/p(\boldsymbol{y}) \propto p(\boldsymbol{y} ,\boldsymbol\tau ,\sigma^2 ,\lambda)$, and maximize the following joint distribution $p(\boldsymbol{y} ,\boldsymbol\tau ,\sigma^2 ,\lambda)$ to get the type-II maximum likelihood estimation of $\boldsymbol\tau$. This joint distribution can be obtained by integrating out $\boldsymbol{w}$ as
\begin{eqnarray*} \nonumber
 p(\boldsymbol{y}, \boldsymbol{\tau}, \sigma^{2}, \lambda)   &= \int p(\boldsymbol{y | w} ,\sigma^{2})p(\boldsymbol{w}|\boldsymbol{\tau}, \sigma^2)p(\boldsymbol{\tau}|\lambda) p(\lambda)p(\sigma^{2})\d\boldsymbol{w} \\
&=  \Big( \frac{1}{2\pi} \Big)^{N/2} |\boldsymbol{C}|^{-\frac{1}{2}}e^{  -\frac{1}{2}\boldsymbol{y}^T\boldsymbol{C}^{-1}\boldsymbol{y}}  p(\boldsymbol{\tau}|\lambda)p(\lambda)p(\sigma^{2}),
\end{eqnarray*}
where  $\boldsymbol{C} = (\sigma^{2} \boldsymbol{I}_N + \boldsymbol{\Phi} \Lambda \boldsymbol{\Phi}^{\top})$. The log of $p(\boldsymbol{y}, \boldsymbol{\tau}, \sigma^{2}, \lambda)$ is given by
\begin{align} \label{llikny}  \nonumber  
 L = &-\frac{1}{2} \log |\boldsymbol{C}| -\frac{1}{2}\boldsymbol{y}^T\boldsymbol{C}^{-1}\boldsymbol{y} + N\log\frac{\lambda}{2}
-\frac{\lambda}{2} \sum_{ i} \tau_i   + a \log b - \log \Gamma (a) \\   &+(a -1) \log \lambda - b \lambda 
+  c \log d - \log \Gamma (c) - (c + 1) \log \sigma^2 - \frac{d} {\sigma^2}.
\end{align}
In the following subsection, an optimisation algorithm for Equation~\eqref{llikny} is presented, and we prove that the algorithm  gives a sparse model because some of the $\tau_i$  will be set to zero. Thus, the corresponding weights and input vectors are pruned from the model according to the construction of the ARD prior in Equation~\eqref{blsw}.

\subsection{Fast optimization algorithm}
A disadvantage of the original RVM method described by \cite{tipping2001sparse} is that it is computationally slow in the maximization of the type-II likelihood.
 The original RVM  begins with all $N$ basis functions included in the model and updates the hyperparameters iteratively. During the updates, some of the basis functions are pruned. However, the first few iterations  require $O(N^3)$ computations. \cite{tipping2003fast} overcome this problem by introducing a Fast Marginal Likelihood Maximization algorithm for Sparse Bayesian Models.  Instead of updating the whole hyperparameter vector $\boldsymbol{\tau}$,  only a single parameter $\tau_i$ is updated at each iteration (step 4 of Algorithm~\ref{alg:seq}). We follow the procedure explained by \cite{tipping2003fast} and choose the $\tau_i$ that gives the largest increase of the log marginal likelihood in Equation~\eqref{llikny}. This fast maximization process is utilized in many sparse learning studies including the work by \cite{ babacan2010bayesian}. 

To find the new value of $\tau_i$, we first rewrite Equation~\eqref{llikny} as:
\begin{eqnarray}  \label{llany} \nonumber 
L =  &-& \frac{1}{2} \Big[ \mathrm{log} | \boldsymbol{C}_{-i} | + \boldsymbol{y}^T\boldsymbol{C}_{-i}^{-1}\boldsymbol{y} + \lambda\sum_{j\neq i}\tau_j \Big]
 + \frac{1}{2} \Big[ \mathrm{log}\frac{1}{1+\sigma^{2}\tau_is_i}+ \frac{q_i^2 \sigma^{2}\tau_i}{1 + \sigma^{2}\tau_i s_i} - \lambda \tau_i \Big]   \\  \nonumber
&+& N\log\frac{\lambda}{2} 
    + a \log b - \log \Gamma (a)  +(a -1) \log \lambda - b \lambda 
+  c \log d - \log \Gamma (c)   \\ &-& (c + 1) \log \sigma^2 - \frac{d} {\sigma^2},
\end{eqnarray}
where \begin{eqnarray} \label{siqi}
s_i = \boldsymbol{\phi}_i^T\boldsymbol{C}_{-i}^{-1}\boldsymbol{\phi}_i, \quad \mathrm{and} \quad q_i = \boldsymbol{\phi}_i^T\boldsymbol{C}_{-i}^{-1}\boldsymbol{y}.
\end{eqnarray}
The notation $\boldsymbol{C}_{-i}$ denotes the covariance matrix $\boldsymbol C$ without the inclusion of the $i$th basis function, $\boldsymbol \phi_i$. The decomposition of the covariance matrix $\boldsymbol C$ into $\boldsymbol{C}_{-i}$ and the other components is explained in Appendix~\ref{ap: covariate}.
The log-likelihood function, $L$, is now decomposed into three parts; the first part is the likelihood where $\tau_i$ and the corresponding $\boldsymbol{\phi}_i$ are excluded, the second part contains the terms that involve $\tau_i$ and $\boldsymbol\phi_i$, while the last part contains all terms not containing $\boldsymbol\tau$.
From this decomposition,
the maximum of $L$ with respect to a single hyperparameter $\tau_i$ is found by taking the partial derivative with respect to $\tau_i$ and setting it equal to zero, which gives
\begin{eqnarray}
\tau_i = \begin{cases}  \frac{-s_i - 2\lambda \sigma^{-2} +   \sqrt{s_i^2 +4q\lambda\sigma^{-2} }}{2\lambda s_i}  & \text{if }    q_i^2 -s_i > \lambda \sigma^{-2} \\
0  & \text{otherwise.} \label{gammaicase}
 \end{cases}
\end{eqnarray}  
The derivation of Equation~\eqref{gammaicase}  can be found in Appendix~\ref{ap: likelihood}. 

After $\tau_i$ is updated we proceed to find the $\lambda$ in Equation~\eqref{gammaicase} that maximizes the log likelihood. By taking the derivative of Equation~\eqref{llikny} with respect to $\lambda$ and setting it to zero we obtain
\begin{equation} \label{estlambda}
{\lambda} = \frac{2(N + a -1)}{\sum_i \tau_i + 2 b}.
\end{equation}
Similarly, a new estimate of $\sigma^2$ is found. When we take the derivative of Equation~\eqref{llikny} with respect to $\sigma^2$, we notice that $\sigma^2$ can be separated from the rest of the components in $\boldsymbol{C}$ such that $\boldsymbol{C} = \sigma^2 \boldsymbol{\tilde{C}}$, where $\boldsymbol{\tilde{C}}$ is independent of $\sigma^2$. The updated value of $\sigma^2$ is then found as
\begin{equation}
{\sigma}^2 = \frac{\boldsymbol{y}^\top \boldsymbol{\tilde{C}}^{-1}\boldsymbol{y} +2 d}{N + 2c + 2}. \label{estsigma}
\end{equation}

In the optimization algorithm, we also have to update the expressions for  $s_i$ and $q_i$. Computing the values of $s_i$ and $q_i$ directly from Equation~\eqref{siqi} requires the inversion of the matrix $\boldsymbol C_{-i}$. Instead, we follow the approach of \cite{tipping2003fast} and calculate: 
\begin{equation}\label{SQbig}
s_i = \frac{S_i}{1-\tau_i \sigma^{2} S_i},  \qquad q_i = \frac{Q_i}{1- \tau_i \sigma^{2} S_i},
\end{equation}
where,
\begin{eqnarray}
S_i &=& \label{SI} \sigma^{-2}\boldsymbol{\phi}_i^\top \boldsymbol{\phi}_i -  \sigma^{-2}\boldsymbol{\phi}_i^\top \boldsymbol{\phi \Sigma \phi}^\top \boldsymbol{\phi}_i\sigma^{-2} ,  \\
Q_i &=& \label{QI} \sigma^{-2}\boldsymbol{\phi}_i^\top \boldsymbol{y} -  \sigma^{-2}\boldsymbol{\phi}_i^\top \boldsymbol{\phi \Sigma \phi}^\top \boldsymbol{ y}\sigma^{-2}.
\end{eqnarray} 
The matrix $\boldsymbol{\Sigma}$ and vector $\boldsymbol{\phi}$ contain only the basis functions that are currently included in the model. 
This computation is therefore much faster than  if we had started with all $N$ basis functions. Algorithm~\ref{alg:seq} summarizes the procedure.
\begin{algorithm}[]
\begin{algorithmic}[1]
 \STATE initialize $\sigma^{2}$ to some sensible value (e.g., var($ \boldsymbol{y} $) $\times$ 0.1)
    \STATE Initialize all $\tau_i = 0$,  $\lambda$ = 0
\WHILE {convergence criteria are not met,}
\STATE Choose a $\tau_i$ 
\IF{$q_i^2 - s_i > \lambda \sigma^{-2}$ and $\tau_i = 0$}
    \STATE Add $\tau_i$ to the model with updated $\tau_i$
  \ELSIF{$q_i^2 - s_i > \lambda \sigma^{-2}$ and $\tau_i > 0$,}
    \STATE Re-estimate $\tau_i$
  \ELSIF{$q_i^2 - s_i < \lambda \sigma^{-2}$,}
    \STATE Set $\tau_i = 0$ 

  \ENDIF
  \STATE Update $\boldsymbol{\Sigma}$ and $\boldsymbol{\mu}$
   \STATE Update $\lambda$ using Equation~\eqref{estlambda}
  \STATE Update $\sigma^2$ using Equation~\eqref{estsigma}  
 \STATE Update $s_i$ and $q_i$ using Equations~\eqref{SQbig} - \eqref{QI}
  \ENDWHILE
\end{algorithmic}
\caption{The Bayesian Lasso Sparse (BLS) Learning Model }
\label{alg:seq}
\end{algorithm}

From  Equation~\eqref{gammaicase} and Algorithm~\ref{alg:seq},  we see that the criteria for setting $\tau_i = 0$ depends on both $\lambda$ and the variance term $\sigma^2$. We will now prove that when $\sigma^2\rightarrow \infty$ then the criteria $q_i^2-s_i \le \lambda\sigma^{-2}$ will be satisfied.
We can write $\boldsymbol{C}_{-i}$ as (see Appendix~\ref{ap: covariate}): 
\begin{eqnarray*}
\boldsymbol{C}_{-i}  &=& \sigma^{2} \boldsymbol{I} + \boldsymbol{\Phi}_{-i} \Lambda_{-i}\boldsymbol{\Phi}_{-i}^{\top} \\
                &=& \sigma^2 \tilde{\boldsymbol{C}}_{-i},
\end{eqnarray*}
where  $\boldsymbol{\Phi}_{-i}$ is the $N \times N$ design matrix where basis function $i$ is removed,  ${ \Lambda}_{-i }$ is the diagonal matrix $\Lambda$  where  the single element $\tau_i$ is removed,  and  $\tilde{\boldsymbol{C}}_{-i}$ denotes $\boldsymbol{C}_{-i}$ where  the component $\sigma^2$ is excluded.
Inserting this decomposition into Equation~\eqref{siqi}, $s_i$ and $q_i$ can be written as
\begin{eqnarray}\label{eq:sisqi_decomposed}
s_i = \sigma^{-2}\tilde{s_i},  \qquad  q_i = \sigma^{-2}\tilde{q_i},
\end{eqnarray}
where $\tilde{s_i} =\boldsymbol{\phi}_i^{\top} \tilde{\boldsymbol{C}}_{-i} ^{-1} \boldsymbol{\phi}_i $ and $\tilde{q_i} = \boldsymbol{\phi}_i^{\top} \tilde{ \boldsymbol{C}}_{-i} \boldsymbol{y}$. The condition in Equation~\eqref{gammaicase} says that
$\tau_i$ is set to zero when $q_i^2-s_i \le \lambda\sigma^{-2}$. Inserting Equation~\eqref{eq:sisqi_decomposed} into this inequality gives
\begin{eqnarray} \nonumber
 \label{siglam}
  \sigma^{-2} \tilde{q_i}^2 -\tilde{s_i}  \leq \lambda.
\end{eqnarray}
We see that as $\sigma^2 \rightarrow \infty$,  the inequality always holds because $\lambda \geq 0$ and $\tilde s_i \geq 0$. Thus, the corresponding $\tau_i$ is set to zero. Therefore, in the BLS method, the information of  $\sigma^2$ is utilized to adjust the number of zero hyperparameters during the estimation of  $\boldsymbol{\tau}$. This feature makes the BLS method more robust to noisy information that might be confused with the real signal information.

\subsection{Prediction}
After the convergence of the learning Algorithm~\ref{alg:seq}, we end up with $A$ nonzero $\tau_i$'s and each of them corresponds to a ``relevance basis function'' and a related ``relevance input vector'' from the training data. For a new input data, $\boldsymbol{x}^*$, we can make predictions based on the posterior of the weights conditioned on  $\boldsymbol{\tau}$ and $\sigma^2$. The predictive distribution \eqref{predy} for the output $y^*$ can  be approximated by
\begin{eqnarray*}  p(y^*|\boldsymbol{y}, \boldsymbol{\tau}, \sigma^2 )  = \int p(y^* | \boldsymbol{w},\boldsymbol\tau , \sigma^2)  p(\boldsymbol{w}|\boldsymbol{y}, \boldsymbol{\tau}, \sigma^2  )   \d\boldsymbol{w} .
\end{eqnarray*}
This distribution is Gaussian with the following predictive mean and predictive variance:
\begin{eqnarray*} 
y^{*} &=& \boldsymbol{\phi}(\boldsymbol{x}^*)^\top \boldsymbol{\mu}, \\ 
\sigma^{*2}&=& \sigma^2 +  \boldsymbol{\phi}(\boldsymbol{x}^*)^\top\boldsymbol{\Sigma}\boldsymbol{\phi}(\boldsymbol{x}^*).
\end{eqnarray*}
 where  $\boldsymbol{\phi}(\boldsymbol{x}^*) =  (1, \phi_{1}(\boldsymbol{x}^*), \dots , \phi_{N}(\boldsymbol{x}^*))^\top$. Note that only the $A$ basis functions corresponding to the nonzero $\tau_i$'s contribute to the posterior mean vector and covariance matrix.

In practice, the predictive mean can be used as a point prediction, and the predictive variance can be used to construct the prediction interval.

\subsection{Relation among BLS, RVM and FLAP}
The BLS method in this paper, together with the RVM \citep{tipping2001sparse} and the FLAP \citep{babacan2010bayesian}, can all be viewed as Sparse Bayesian Learning methods with a hierarchical structure. The main difference of these three methods lies at the choice of prior distribution for the weight parameters.
The  RVM places a zero-mean Gaussian prior on the weight parameters  in Equation~\eqref{linmod}, where each weight parameter has its own precision parameter $\alpha_i$. These hyperparameters have their own hyperprior distribution which is a Gamma distribution with parameters $a$ and $b$: 
\begin{eqnarray*} \begin{split}
p(\boldsymbol{w}| \boldsymbol\alpha) &= \prod_{i=0}^N \mathcal{N} (w_i | 0, \alpha_i^{-1} ),  \hspace{1.2cm} \alpha_i \geq 0, \\
p(\boldsymbol{\alpha}) &= \prod_{i=0}^N \text{Gamma}(\alpha_i | a, b), \qquad a,b \geq 0. \end{split} 
\end{eqnarray*}
Based on this structure, \cite{tipping2001sparse} shows that the underlying marginal prior, $p(\boldsymbol{w})$, is a Student's t-distribution. 

The FLAP method by  \cite{babacan2010bayesian} 
 also use a Gaussian prior on $\boldsymbol{w}$, where the hyperparameters in $\boldsymbol{\tau}$ are defined directly as the variances of the weight parameters, and an exponential hyperprior is set to those hyperparameters. 
 The basic prior for $\boldsymbol{w}$ after integrating all the hyperparameters in $\boldsymbol{\tau}$ is a Laplace distribution:
\begin{eqnarray*}
 p(\boldsymbol{w}| \lambda) = \prod_{i=0}^N \frac{\sqrt{\lambda}}{2} e^{-\sqrt{\lambda} |w_i|}, \qquad \lambda \geq 0. 
\end{eqnarray*} 

Thus, the idea of the sparse setting in the BLS method is similar to the Fast Laplace method described by  \cite{ babacan2010bayesian}. The main difference is that, by using the
conditional prior in Equation~\eqref{bpriorw2} for the weights in the BLS model,  the criteria for letting $\tau_i = 0$ will now also depend on $\sigma^{2}$. As $\sigma^{2}$ is a measurement of the extent of the noise in the dataset, we expect that the BLS method will be more robust to the data noise.

\section{BLS for nonlinear regression}  \label{sec:results}

In this section, we  compare the BLS with the RVM~\citep{tipping2003fast} and the FLAP~\citep{babacan2010bayesian}. We use simulated datasets to be able to compare the estimated $\sigma^2$ to the true vale. For all the methods, we use the Gaussian kernel
$K(\boldsymbol x_i, \boldsymbol x_j) = \mathrm{exp}(-r^2 \| \boldsymbol x_i - \boldsymbol x_j \| ),$
where the kernel parameter $r$ is determined by using five-fold cross validation.

For the Gamma priors in Equations~\eqref{priorlasso} and \eqref{priorsigma} a common practice is to set them to zero in order to obtain uniform hyperpriors for $\lambda$ and $\sigma^2$. In our implementation of the BLS model we therefore set $a=b=c=d=0$. The RVM method uses the same procedure to obtain a uniform hyperprior for $\sigma^2$, but the model does not include $\lambda$. 
In the numerical results of~\cite{babacan2010bayesian}, they use a uniform hyperprior for $\lambda$ in the FLAP method. However, they use a fixed value for $\sigma^2$. This is done due to the under-determined nature of the compressive sensing problem that they apply the method to, which makes the estimates of $\sigma^2$ unstable in the early iterations. In the results of this paper we follow the same setup as in~\cite{babacan2010bayesian} and use $\sigma^{-2} = 0.01 \| \boldsymbol y\|_2^2$. In addition, we include a version of the FLAP method where $\sigma^2$ is estimated in the same manner as the BLS and the RVM. In order to distinguish this approach from the original FLAP, we denote it as FLAP$_\sigma$ in the sections below. We include the results of both FLAP and FLAP$_\sigma$, because, to our awareness this is the first time the FLAP method is tested on general nonlinear regression problems.


\subsection{  The Sinc function} \label{sec:sinc_func}
We first consider the Sinc function, $f(x)= \sin(x)/x$,  a  benchmark function that is frequently used to evaluate how kernel-based learning methods perform~\citep{vapnik1997support, tipping2001sparse, schmolck2007smooth}. 
We use the same procedure as \cite{tipping2001sparse} where the model is built based on 100 training data while the error is calculated with respect to the true function by using 1000 test data. 

\begin{figure}[]
\centering
\includegraphics[width = .99\textwidth]{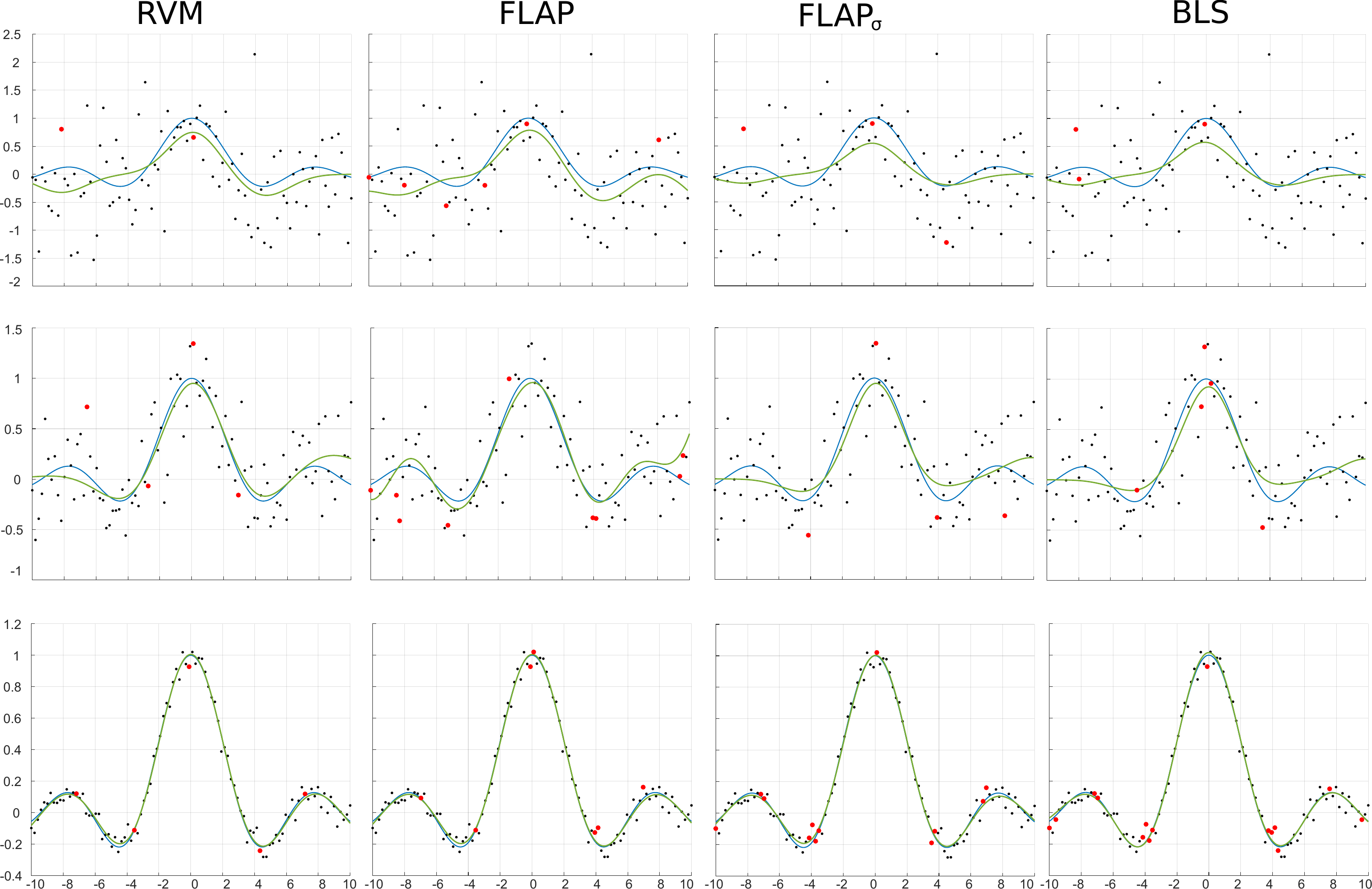}  
\caption{ The Sinc function (blue line) and its prediction (green line) from the data  generated with $\sigma = 0.05, 0.3$ and $0.7$ from top to bottom row. The red dots are the relevance vectors and the black dots are the remaining training data.
}
\label{fig:sinc}
\end{figure}
Figure~\ref{fig:sinc} shows the  results of the four methods on datasets with standard deviation $\sigma$~=~0.05, 0.3 and 0.7. The black dots represent the training data from the model ${y}= f(x)+ \epsilon$ where $x \in [-10,10]$.  The same data set is used for all methods to better compare them. The location of the nonzero weighted input vectors, often referred to as the relevance vectors, are represented by the red circles. The blue lines correspond to the Sinc function, $f(x)$, while the green lines are the prediction of each method applied to the test data. 
From Figure~\ref{fig:sinc} we observe that when $\sigma=0.05$, the approximations of all four methods perform well and almost overlap with the true function.  When $\sigma$ is set to 0.3, the methods can still capture the general form of the original function, except at the boundaries where they produce a more rough approximation. However, when  $\sigma$ is set to 0.7, the training data begin to loose the original shape of the Sinc function, and the approximations of all four methods have issues capturing the Sinc function shape.

To measure the prediction accuracy of the methods, we generate 100 datasets for $\sigma$~=~0.05, 0.1, 0.3, 0.5 and 0.7 and run each method on these datasets totalling $100\times 5\times 4 = 2000$ runs.
For each value of $\sigma$, method and dataset, we calculate the root mean squared error, number of relevance vectors and the estimated standard deviation.
Table~\ref{tab:1dsinc} shows the average values of these quantities, denoted by RMSE, NOV and $\hat \sigma$. When the standard deviation is small ($\sigma\le 0.3$), all four methods give good approximations of the true function with low RMSE. For higher standard deviations the RMSE is higher but comparable for all methods. 
We observe that the NOV of all methods except FLAP is decreasing as the standard deviation increases. Thus, the methods that estimate the standard deviation pick fewer relevance vectors when the noise is large. The RVM and the BLS give the most accurate estimates of the standard deviation. In all, Table \ref{tab:1dsinc} shows that the BLS in general perform as well as the established methods  RVM and FLAP on the Sinc function benchmark.

\input{tabel/sinc1.tex}

\subsection{The Bump dataset} 
In this section we investigate how the methods perform on a dataset with high frequency and spikes.  The so called Bump function from~\cite{donoho1994ideal} is used to simulate data with sample size $N=120$ and different signal to noise ratios (SNR).

\begin{figure}[]
\centering
\includegraphics[width = .98\textwidth]{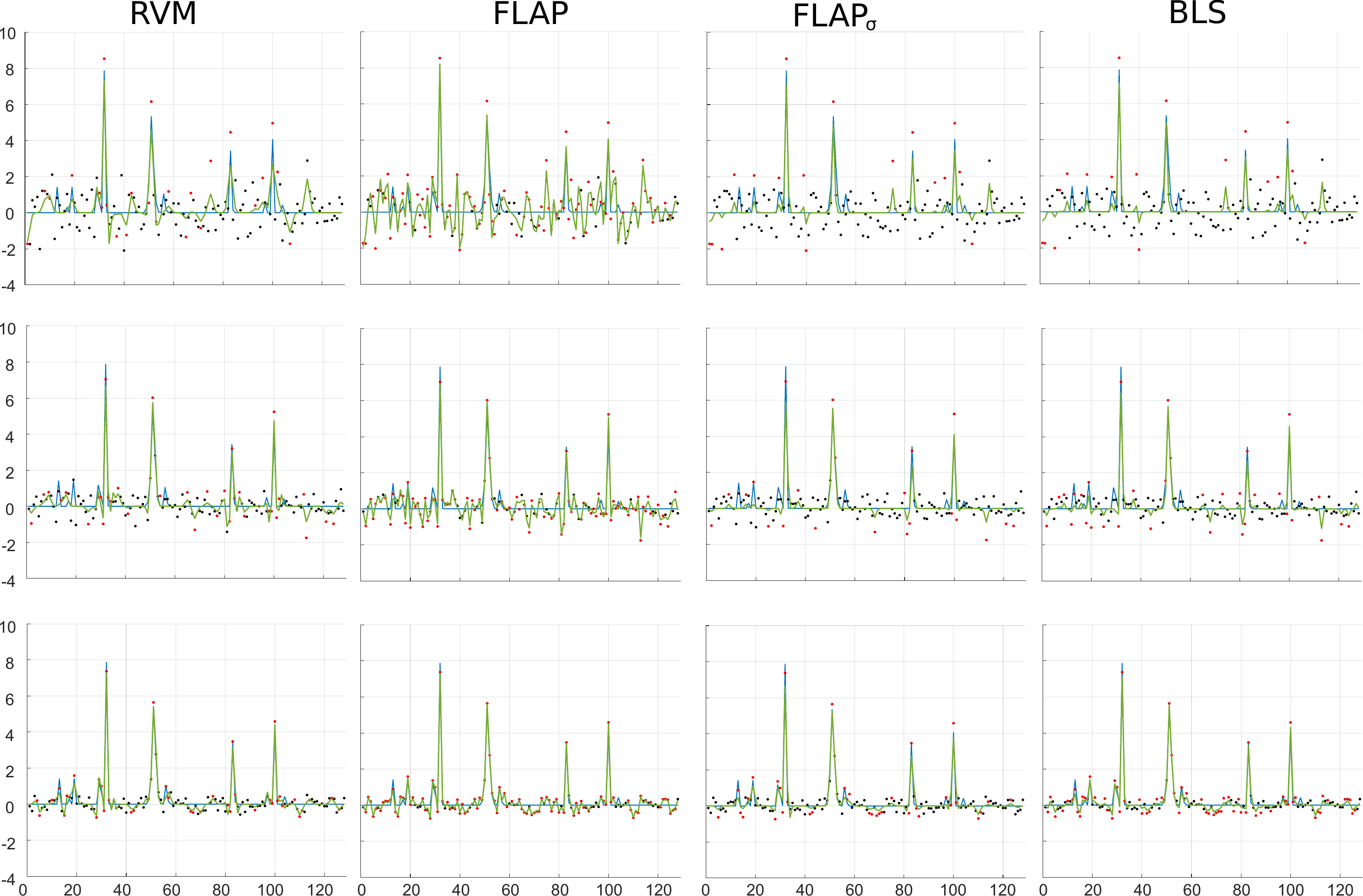} 
\caption{ The Bump function (blue line) and its reconstruction (green line) from the data that are generated for different values of SNR = 1, 3 and 10 (from top to bottom row). The red dots are the relevance vectors and the black dots are the remaining data.}
\label{fig:bump}
\end{figure}

Figure~\ref{fig:bump} shows the predictions of the four methods when SNR~$= 1, 3$ and $ 10$. In this figure, the true Bump function is represented by the blue lines and the reconstruction from the noisy data is the green lines. The red dots represent the relevance vectors and the black dots represent the remaining data samples.  The same data is used for all methods to compare them.
When the data is less noisy, with SNR $= 10$,  all four methods can capture the bump signal well, however, when the noise is larger, the FLAP tends to overfit the data compared to the other methods.  When SNR~$=1$, it is the BLS and FLAP$_\sigma$ that capture the signal best.

As for the Sinc function, we generate $100$ random datasets for SNR = 1, 2, 3, 4, 5 and 10, and utilize RMSE, NOV and $\hat\sigma$ to evaluate the four methods. The results are presented in Table~\ref{tab:1dbump}. We observe that the predictions of the BLS method have the lowest RMSE of all four methods, especially when the dataset is noisy (SNR~$\le 2$). The predictions of the FLAP$_{\sigma}$ have a lower RMSE than RVM and FLAP, except for SNR~$=10$.  Table~\ref{tab:1dbump} also shows that the NOV for the FLAP is roughly twice that of the other methods. This is consistent with the indication from Figure~\ref{fig:bump} that FLAP might overfit the data. The NOV values for the other models are more similar, but we notice that the BLS is the most sparse model when the SNR is low and the dataset is noisy. The NOV of the methods that estimate $\sigma^2$ decreases as the noise of the dataset increases, similar to the results in Section~\ref{sec:sinc_func}.

\input{tabel/bump.tex}

Compared to the smooth Sinc function where  all  four methods gave similar results, the BLS shows better result, both in terms of RMSE and NOV, than the other three methods for the irregular and noisy Bump dataset. 
 
\newpage

\section{BLS for variable selection} \label{sec:variablesel}
Sections~\ref{sec:bls} and~\ref{sec:results} derive and test
the kernel-based  framework where general nonlinear regression problems can be solved.  
This section describes how the BLS method can perform  variable selection in  multiple linear regression models.
Consider the linear regression problem 
\begin{equation} \label{BLReg}
  \boldsymbol{y} = \boldsymbol{ X }\boldsymbol{\beta} + \boldsymbol{\epsilon} ,   \qquad \boldsymbol{\epsilon}\sim \mathcal{N}( 0, \sigma^2\boldsymbol{I}_N),
\end{equation}
where $\boldsymbol{\beta} = (\beta_1, \dots, \beta_P )^\top$ is a vector that holds  regression coefficients, $\boldsymbol{y}$ is the response vector,  and $\boldsymbol{ X } $ is the $N\times P $ matrix of standardized variables. The learning algorithm of the BLS method is generally the same as in Section~\ref{sec:bls}, and the main difference is that all the $\boldsymbol{\Phi} $ matrices will be $\boldsymbol{ X } $ matrices and no kernel function is needed.

One of the main features of the Lasso by \cite{tibshirani1996regression} is its ability to do variable selection. On the other hand, the Bayesian Lasso by~\cite{park2008bayesian} does not perform variable selection because the regression coefficients in $\boldsymbol{\beta}$ are estimated by the Gibbs sampling. This motivates the use of the BLS method where variables related to pruned coefficients will be deleted and sparsity is achieved in variable domain. 

For the Gamma priors in Equations~\eqref{priorlasso} and \eqref{priorsigma}, we set $a=b=c=d=0$ for all the results in this section, similar as done in Section~\ref{sec:results}.



\subsection{Simulated data}

This sub-section compares the BLS for variable selection to the Bayesian Lasso~\citep{park2008bayesian} and the Lasso~\citep{tibshirani1996regression} using simulated data.  
   The data are simulated from 
\begin{eqnarray*} 
y_i = \boldsymbol x_i^\top \boldsymbol \beta + \epsilon_i, \qquad i=1, \dots,N,
\end{eqnarray*}
where $\epsilon_i \sim N(0, \sigma^2)$.  We consider three different simulation studies where we run 100 simulations for each study. In Simulation 1,  the number of observations is much larger than the number of variables. In Simulation 2, the number of observations is slightly higher than the number of variables, while in Simulation 3, the number of observations is less than the number of variables.

For all studies  we generate a training set of size 50, while the test error is calculated with respect to the true model by using a test set of size 100. For the Bayesian Lasso, the posterior means are calculated based on 15 000 samples after discarding 3000 burn-in samples. Further, the  hyperparameters $a$ and $b$ in the Gamma prior for $\lambda^2$  is set to $0.1$.  For the Lasso model, the tuning parameter $\lambda$ is selected by using 10-fold cross validation.

In the first simulation study we set $\boldsymbol \beta = (3, 1.5, 0, 0, 2, 0, 0, 0)^\top$  and consider three different scenarios setting $\sigma = 1, 3$ and $5$. 
The pairwise correlation between $\boldsymbol x_i$ and $\boldsymbol x_j$, $i \neq j$, is $\rho^{|i-j|}$ with $\rho = 0.5$.
A similar example was used in the original paper by \cite{tibshirani1996regression} and in several later studies \citep{leng2006note,li2010bayesian, alhamzawi2018bayesian}. 

The simulation results are reported in Table~\ref{tab: sim_lasso_selected}, where the frequency of selected coefficients from the 100 simulations are listed. As  the true values of $\beta_1, \beta_2$ and $\beta_5$ are nonzero, a correct model should select these three as the nonzero coefficients. The table shows that both the Lasso and the BLS selects these coefficients in all 100 repetitions when $\sigma=1$, while for $\sigma=3$ and $5$, the Lasso has the highest frequency. The Bayesian Lasso selects all coefficients in every repetition, which is as expected. Maybe more interesting, is that the BLS selects the remaining coefficients, whose true values are zero, less frequently than the Lasso.

Table~\ref{tab: sim_lasso_selected} also reports the average number of nonzero coefficients (NOC) along with the average RMSE and corresponding sample standard deviations. The NOC values from the Bayesian Lasso are 8 in all three cases, which means that none of the estimated coefficients are  zero, and the NOC values from the BLS are closer to 3 than the NOC values of the Lasso.  
The BLS model has the lowest RMSE for all cases of $\sigma$. Thus, Table~\ref{tab: sim_lasso_selected}  shows that the BLS provides on average the most sparse estimated model with the lowest prediction error.

\begin{table}[]
\centering
\caption{Results from Simulation 1. For each value of the true standard deviation $\sigma$, 100 datasets are generated. The table reports the average results of the  BLS, the Bayesian Lasso (BL) and the Lasso. The columns $\beta_1,\ldots,\beta_8$ list the number of times each coefficient is selected. The average number of nonzero coefficients, NOC, the average root mean square error, RMSE, and the average of estimated standard deviation, $\hat{\sigma}$, are listed  along with the corresponding sample standard deviation SD.
   }  \vspace{0.2cm} 
\label{tab: sim_lasso_selected}
\def\arraystretch{1.9}
\scalebox{0.9}{

\begin{tabular}{llllllllllllllll}
   \hline
$\sigma$  & Method & $\beta_1$ & $\beta_2$ & $\beta_3$ &$\beta_4$ & $\beta_5$ & $\beta_6$ & $\beta_7$ & $\beta_8$ \hspace{0.2cm} & NOC (SD) & RMSE (SD) \\ 
  \hline
1   & BLS & 100 & 100 & 34 & 36 & 100 & 31 & 24 & 20   & 4.5 \ (0.95) & 0.298 \ (0.13)  \\
 & BL  & 100  &  100  &   100 & 100 & 100 & 100 & 100 & 100   & 8 \ (0) & 0.412 \ (0.12)  \\
  & Lasso  & 100 & 100 & 53 & 50 & 100 & 51 & 43 & 49   & 5.5 \ (1.45) & 0.397 \ (0.14)\\
 
   \hline
   3  & BLS &  100 & 96 & 30 & 29 & 99 & 28 & 28 & 22  &  4.3 \ (0.98) &  0.984 \ (0.39)\\
 & BL & 100   & 100   & 100 & 100 & 100 & 100 & 100 & 100  & 8 \ (0) & 1.210 \ (0.37) \\
 & Lasso & 100 & 98 & 41 & 56 & 100 & 40 & 45 & 44  & 5.2 \ (1.60) & 1.204 \ (0.44) \\
  \hline
   5  & BLS & 100 & 83  & 19  & 20 & 80 & 19 & 14 & 17  & 3.5  \ (0.87) & 1.788 \ (0.72)  \\
 & BL  & 100 & 100  & 100 & 100 & 100 & 100 & 100 & 100 & 8 \ (0) &  1.892 \ (0.59)  \\
 & Lasso & 99 & 82 & 38 & 40 & 93 & 37 & 39 & 35  & 4.6 \ (1.59) & 1.960 \ (0.61) \\
   \hline
\end{tabular}
}
\end{table}

The average of the estimated values of $\boldsymbol \beta$ and the corresponding standard deviation is presented in Table~\ref{tab: sim_lasso_coef}. All methods give similar values for the coefficients, both for the nonzero and zero coefficients. We observe that while Bayesian Lasso never sets a coefficient to zero, the values of these coefficients are close to zero.
\begin{table}[]
\centering
\caption{The average coefficient estimates and corresponding sample standard deviations based on 100 datasets for  Simulation 1.  }  \vspace{0.2cm} 
\label{tab: sim_lasso_coef}
\def\arraystretch{1.9}
\scalebox{0.8}{
\hspace{-1,5cm}
\begin{tabular}{llllllllllll}
   \hline
 $\sigma$  & Method & $\beta_1=3$ & $\beta_2=1.5$ & $\beta_3=0$ &$\beta_4=0$ & $\beta_5=2$ & $\beta_6=0$ & $\beta_7=0$ & $\beta_8=0$ \hspace{0.2cm} \\ 
  \hline
1   & BLS &  2.988 \ (0.17) &  1.450 \ (0.16)  &  -0.001 \ (0.10) & 0.021 \ (0.12)  &  1.958 \ (0.18) & 0.005 \ (0.10)  &  -0.013 \ (0.09) & -0.002 \ (0.07) \\
  & BL  &  2.986 \ (0.18) & 1.483 \ (0.20) & 0.001 \ (0.17) & 0.005 \ (0.15) & 1.972 \ (0.19) & 0.032 \ (0.16) & -0.023 \ (0.15) & 0.017  \ (0.15)  \\
  & Lasso  &  2.932 \ (0.20) & 1.440 \ (0.17) & 0.026 \ (0.12) & 0.020 \ (0.14) & 1.896 \ (0.17) & 0.028 \ (0.12) & -0.004 \ (0.14) & -0.004 \ (0.11) \\
 
   \hline
   3  & BLS & 2.896 \ (0.54) &  1.128 \ (0.62)  &  0.098 \ (0.26) & 0.112 \ (0.30) &  1.640 \ (0.50)  &  0.067  \ (0.26) & 0.016 \ (0.18)  &  -0.014  \ (0.22) \\
 & BL &  2.859 \ (0.55) & 1.209 \ (0.61) & 0.079 \ (0.42) & 0.133 \ (0.37) & 1.567 \ (0.51) & 0.111 \ (0.35) & 0.028 \ (0.31) & 0.008 \ (0.31) \\
 & Lasso & 2.784 \ (0.57) & 1.260 \ (0.56) & 0.098 \ (0.30) & 0.079 \ (0.38) & 1.706 \ (0.52) & 0.006 \ (0.45) & 0.073 \ (0.37) & 0.019 \ (0.36)    \\
  \hline
   5  & BLS & 2.597 \ (0.96)  & 1.069 \ (0.85)  & 0.138 \ (0.35) & 0.145 \ (0.35) & 1.051 \ (0.83) & 0.087 \ (0.27) & -0.026 \ (0.21) & -0.017 \ (0.25)   \\
 & BL  &  2.507 \ (0.80) & 1.171 \ (0.71) &  0.191 \ (0.49) & 0.240 \ (0.61) & 1.308 \ (0.72) & 0.193 \ (0.50) & 0.045 \ (0.43) & -0.042 \ (0.52) \\
 & Lasso &  2.606 \ (0.93) & 1.135 \ (0.92) & 0.030 \ (0.52) & 0.142 \ (0.63) & 1.485 \ (0.89) & 0.184 \ (0.58) & -0.030 \ (0.55) & 0.096 \ (0.53) \\
   \hline
\end{tabular}
}
\end{table}

In the second simulation study, we simulate data where the dimension $P$ of the input variables is close to the sample size $N$. We simulate a dataset with 50 observations for the training set and with 40 predictors. We set
\begin{eqnarray*}
 \boldsymbol \beta = (\underbrace{3,\dots,3}_{5},\underbrace{3,\dots,3}_{5},\underbrace{3,\dots,3}_{5},\underbrace{0,\dots,0}_{25})^\top,
\end{eqnarray*}
and $\sigma =1$.  For the variables in $\boldsymbol X$  we first generate $Z_1, Z_2$ and $Z_3$ independently from $N(0,1)$. Then let
\begin{eqnarray*}
 x_i &=& Z_1 + e_i, \qquad  \text{for } i=1,\dots,5,\\
 x_i &=&Z_2 + e_i,\qquad \text{for } i=6,\dots,10,\\
 x_i &=&Z_3 + e_i, \qquad \text{for } i=11,\dots, 15,
\end{eqnarray*}
where $e_i \sim N(0, 0.01)$ for $ i=1,\dots,15$. For the remaining 25 variables, $i=16,\dots,40$, we set $x_i \sim N(0,1)$. The results from Simulation 2 are presented in Table~\ref{tab: sim_largeM}. For Simulation 2 the Lasso model barely obtains the lowest RMSE value before BLS. The Bayesian Lasso RMSE is significantly higher. A correct model should select the 15 nonzero coefficients and set the remaining 25 to zero. Table~\ref{tab: sim_largeM} also shows that the average NOC value from both the Lasso model and the BLS method overestimate the number of coefficient that are different from zero.
\begin{table}[]
\centering
\caption{Result for Simulation~2 and 3, where average results for 100 datasets are reported.  The average number of nonzero coefficients,  NOC, and  the average root mean square error, RMSE, are listed  along with the corresponding sample standard deviation SD.  }  \vspace{0.2cm} 
\label{tab: sim_largeM}
\def\arraystretch{1.9}
\scalebox{0.9}{

\begin{tabular}{lllllll}
  \hline
 & & & Method & NOC (SD) & RMSE (SD) &  \\ 
  \hline
Simulation 2 & &  & BLS & 24.5 \ (2.83) & 0.947 \ (0.18) \\
 & & & BL &   40 \ (0)  &  1.569 \ (0.29) \\
 & & & Lasso & 22.7 \ (5.10) & 0.943 \ (0.21) \\
 
   \hline
Simulation 3    & & & BLS & 47.2 \ (2.14) & 2.315 \ (0.44) \\
& & & BL & 60 \ (0) & 30.84 \ (1.96)  \\
& & & Lasso &  49.1 \ (2.76)   & 2.871  \ (0.39)\\
   \hline
   
\end{tabular}
}
\end{table}

In Simulation 3 we simulate a training dataset with 50 observations and with 60 predictors. We set 
\begin{eqnarray*}
 \boldsymbol \beta = (\underbrace{5,\dots,5}_{10},\underbrace{3,\dots,3}_{10},\underbrace{3,\dots,3}_{10},\underbrace{2,\dots,2}_{10},\underbrace{2,\dots,2}_{10},\underbrace{0,\dots,0}_{10})^\top.
\end{eqnarray*}
Similar to Simulation 2, for each group of ten variables we generate $Z_i~\sim N(0,1)$ and set
\begin{align*}
 x_i &= Z_1 + e_i, \qquad  \text{for } i=1,\dots,10,
 \qquad  &&x_i = Z_2 + e_i, \qquad  \text{for } i=11,\dots,20,\\
 x_i &=Z_3 + e_i,\qquad \text{for } i=21,\dots,30,
 \qquad  &&x_i = Z_4 + e_i, \qquad  \text{for } i=31,\dots,40,\\
 x_i &=Z_5 + e_i, \qquad \text{for } i=41,\dots, 50,
\end{align*}
with $e_i \sim N(0, 0.01)$. For the remaining 10 variables we set $x_i \sim N(0, 1)$.
From Table~\ref{tab: sim_largeM} we see that the BLS has the lowest average RMSE value for Simulation 3, slightly lower than the Lasso. The Bayesian Lasso does not perform well when $P>N$ on this dataset, however, we have observed that the RMSE drastically improves when the number of samples increases.


\subsection{The Diabetes data}

The final simulation study uses the diabetes data presented by~\cite{efron2004least}. The dataset was also used in a study  by \cite{park2008bayesian}  to compare the  performance of the Bayesian Lasso with the Lasso and the  Ridge Regression. 
 The response is a measure of disease progression of 442 patients measured by 10 variables, one year after baseline. 
We standardize the predictors to have zero mean and unit variance. 
Table~\ref{tab:diabetes} compares the estimates from the BLS, the Bayesian Lasso and the Lasso. For the BLS,  the point estimates in $\boldsymbol{\beta}$ are the mean of the posterior distribution. For the Bayesian Lasso, we use the same settings as  \cite{park2008bayesian} and  report the  posterior median estimates obtained by using the Gibbs sampler. 
  The Bayesian $95 \% $ credible intervals are also given.  The BLS estimates are within the credible intervals of the Bayesian Lasso for all coefficients. We also notice that the BLS and the Lasso set the same coefficients to zero. 
\input{tabel/diabetes.tex}%

To compare the prediction performance, the dataset is randomly split in two parts, $70~\%$ of the data is used as a training set and the remaining $30~\%$ is used as a test set. We carry out 100 repetitions and report the average test RMSE and the average number of selected coefficients (NOC) with corresponding sample standard deviation~(SD). For the Lasso, we used 10-fold cross-validation to select the value of $\lambda $.  The results given in Table~\ref{tab:DBmse} shows that for this empirical example, although the RMSE is more
or less the same for the three methods, BLS is most sparse with lowest number of
selected variables.

\begin{table}[]
\centering
\caption{Results of the Diabetes data. The data is split into $70~\%$ training data and $30~\%$ test data for 100 repetitions, where the average test root mean square error, RMSE, and average number of coefficients, NOC, are listed along with the corresponding sample standard deviations, SD.   }  \vspace{0.2cm} 
\label{tab:DBmse}
\def\arraystretch{1.2} 
\begin{tabular}{llllll}
  \hline
 Models &  &  & RMSE (SD)   & & NOC (SD) \\ 
  \hline 
  BLS   & & & 55.10 (2.64)  & &  6.35 (0.73) \\  
  BL  & & &  55.06 (2.67)  &  & 10  (0) \\  
  Lasso   &  & & 55.11 (2.69)  &  & 7.97 (1.26)  \\  
   \hline
\end{tabular}
\end{table}

%
%
%
%
%
%
%
%

\section{Conclusion } \label{sec:conclusion}
In this paper a new sparse Bayesian learning method, called the Bayesian Lasso Sparse (BLS) method is presented. 
The main features of the BLS can be summarized in three points: 
(I) The developed BLS method extends the Bayesian Lasso by \cite{park2008bayesian} to deal with general nonlinear supervised learning problems, with the help of the kernel-based framework in \cite{tipping2001sparse} and the fast learning process in  \cite{tipping2003fast}. The resulting method is a sparse Bayesian method that is shown to achieve sparsity in the sample domain and good predictive properties.
(II) The  prior  distribution  of  BLS  is  conditioned on the variance of random noise, and the BLS is shown to be robust to the irregular datasets and high variance. We analyse how the  posterior  estimation  of  the  weight parameters is adjusted by the variance of the noise. 
(III) We present how the BLS can be used in multiple linear regression. The BLS can here achieve variable selection automatically by a pure data-driven process.




The developed BLS method is compared to the well-known methods RVM, FLAP, BL, and Lasso in addition to FLAP$_\sigma$. To investigate the performance of the methods, we carry out a comprehensive study with both simulated and real data. For the simulated datasets the methods are exposed to various extents of noise. For the nonlinear regression test cases included in this paper, the FLAP$_\sigma$ method mostly performs better than FLAP. We do not observe that the FLAP$_\sigma$ method is unstable at early iterations, which was observed for the compressive sensing reported by~\cite{babacan2010bayesian}. Our results show that the performance of the BLS is compatible with the other methods, with low test error, few “relevance" vectors and low bias estimation of ${\sigma^2}$.



\begin{appendices}

\section{}\label{ap: covariate}
In order to obtain the  expression for the log likelihood in Equation~\eqref{llany}, we  decompose the covariance matrix in the log-likelihood in Equation~\eqref{llikny}  as:

\begin{eqnarray} \nonumber 
\boldsymbol{C}  &=& \sigma^{2} \boldsymbol{I} + \sum_{m \neq i}  \sigma^{2}\tau_m \boldsymbol{\phi}_m \boldsymbol{\phi}_m^T +  \sigma^{2}\tau_i\boldsymbol{\phi}_i \boldsymbol{\phi}_i^T \\
&=& \boldsymbol{C}_{-i} +  \sigma^{2}\tau_i\boldsymbol{\phi}_i \boldsymbol{\phi}_i^T,  \label{nyexsig}
\end{eqnarray}
where $\boldsymbol{C}_{-i}$ denotes $\boldsymbol{C}$ without the inclusion   of basis function $i$. We next use the  Woodbury identity on the expression for the covariance matrix  in Equation~\eqref{nyexsig}, such that the inverse of the covariance matrix is written as:
\begin{eqnarray*}
\boldsymbol{C}^{-1} = \boldsymbol{C}_{-i}^{-1} - \frac{\boldsymbol{C}_{-i}^{-1} \boldsymbol{\phi}_i \boldsymbol{\phi}_i^{T}\boldsymbol{C}_{-i}^{-1}}{\sigma^{-2} \tau_i^{-1} + \boldsymbol{\phi}_i^{T}\boldsymbol{C}_{-i}^{-1} \boldsymbol{\phi}_i}.
\end{eqnarray*}
Finally we use  the determinant identity  to obtain the  decomposition of the determinant: 
\begin{equation*}
|\boldsymbol{C}| = |\boldsymbol{C}_{-i}| \  |1 + \sigma^{2} \tau_i\boldsymbol{\phi}_i^{T}\boldsymbol{C}_{-i}^{-1} \boldsymbol{\phi}_i |.
\end{equation*}
These last two  expressions can be inserted in  Equation~\eqref{llikny}, which results in Equation~\eqref{llany}.

\section{}\label{ap: likelihood}
From the decomposition of the log likelihood given in Equation~\eqref{llany}, we can find the derivative of  $L$ with respect to $\tau_i$,  where the other parameters are considered as fixed. 

\begin{align*} \frac{\d L}{\d \tau_i}  
 &= \frac{1}{2} \bigg[ -\frac{s_i}{\sigma^{-2} + \tau_i s_i}  + \frac{q_i^2 \sigma^{-2}}{(\sigma^{-2} + \tau_i s_i)^2} -\lambda \bigg] \\
&=- \frac{(\tau_i^2 \kappa_1 + \tau_i \kappa_2 + \kappa_3 ) }{2(\sigma^{-2} + \tau_i s_i)^2}  , 
\end{align*}
where $\kappa_1 = \lambda s_i^2,$ $ \kappa_2 = s_i^2 +2 s_i \lambda \sigma^{-2}  $ and $\kappa_3 = \sigma^{-2}(\lambda \sigma^{-2} + s_i-q_i^2)$.  
The numerator has a quadratic form while the denominator is always positive so that $d  L/d \tau_i$ $ = 0$ is satisfied at
\begin{eqnarray}
\tau_i  &=&  \label{derivgamma}
 \frac{-s_i(s_i + 2 \lambda \sigma^{-2}) \pm s_i \sqrt{\Theta}}{2\lambda s_i^2}, 
\end{eqnarray}
where $\Theta = (s_i + 2  \lambda \sigma^{-2} )^2 -4\lambda  \sigma^{-2}(\lambda \sigma^{-2} -( q_i^2-s_i))$. 
By analysing the terms we see that if  $q_i^2-s_i < \lambda \sigma^{-2}$, then $\Theta^2< s_i +2\lambda \sigma^{-2}$, and both solutions  of Equation~\eqref{derivgamma} are negative. Furthermore, since $\d L/\d\tau_i$ evaluated at $\tau_i = 0$ is negative, the maximum occurs at $\tau_i = 0$.  In the other situation, when    $q_i^2 -s_i > \lambda \sigma^{-2}$, there are two real solutions of Equation~\eqref{derivgamma}, one negative and one positive. The positive solution from Equation~\eqref{derivgamma}  maximizes $L$ since $ \d L/\d\tau_i$ is positive when evaluated at $\tau_i = 0$  and negative at $\tau_i = \infty$. The maximum of $L$, when holding the remaining components fixed, is therefore obtained at:

\begin{align*}
\tau_i = \begin{cases}  \frac{-s_i(s_i + 2 \lambda \sigma^{-2}) +  s_i\sqrt{\Theta}}{2\lambda s_i^2}  & \text{if }    q_i^2 -s_i > \lambda \sigma^{-2} \\
0  & \text{otherwise.} 
 \end{cases}
\end{align*}  
Notice that the expression for $\Theta$ can be simplified to $\Theta =  s_i^2 +4q\lambda\sigma^{-2}$ which results in Equation~\eqref{gammaicase}.

\end{appendices}

\bibliographystyle{apa}   
\bibliography{BLS}   

\end{document}

%% file: tabel/sinc1.tex

\begin{table}[]
\centering
\caption{Results of the simulation study for the Sinc function. For each value of $\sigma$, 100 datasets were generated. The average number of relevance vectors, NOV, the average root mean square error, RMSE, and the average of estimated noise standard deviation, $\hat{\sigma}$, are reported for each method along with the corresponding sample standard deviation, SD, in the parenthesis. 
}  \vspace{0.2cm} 
\label{tab:1dsinc}
\def\arraystretch{1.9}
\scalebox{0.75}{
\begin{tabular}{llllll}
  \hline
True value of $\sigma$ & &  Method \hspace{0.1cm} & NOV (SD) \hspace{0.1cm}  & RMSE (SD) \hspace{0.1cm} & $\hat{\sigma}$ (SD) \\ 
  \hline
 0.05  & &  RVM & 5.23 \ (0.45)  &  0.016 \ (0.003)   &   0.049 \ (0.004)   \\
     & & FLAP & 6.53 \ (1.21)   &   0.015 \ (0.003)    & -\\
  & & FLAP$_{\sigma}$ &   12.30 \ (1.79)  &  0.017 \ (0.004)   &   0.047  \ (0.004)   \\
  & & BLS & 15.23 \ (3.12) &  0.018 \ (0.004)   &       0.049 \ (0.005)      \\
   \hline
  0.1 & &  RVM & 5.13 \ (0.65)  &  0.034 \ (0.008)    &   0.092 \ (0.008)  \\
 & & FLAP &  7.57 \ (1.74)   &  0.029 \ (0.006)     & - \\
 & & FLAP$_{\sigma}$ &  19.01  \  (3.88)   &   0.039  \  (0.008)  &   0.079  \ (0.006) \\
 & & BLS & 17.79 \ (3.11)   &   0.039 \ (0.008) &  0.096 \ (0.010)   \\
  \hline
  0.3 & &  RVM &  4.23 \ (0.83)   &  0.092 \ (0.019)  &   0.30 \ (0.022)\\
 & & FLAP & 8.80 \ (2.18) &  0.088 \ (0.021)  & - \\
 & & FLAP$_{\sigma}$ & 5.41 \ (1.29) & 0.096  \ (0.017) &     0.25  \  (0.041) \\
 & & BLS & 5.39 \ (1.45) & 0.099 \ (0.016) &  0.30 \ (0.029) \\
  \hline
  0.5 & &  RVM &  3.64 \ (1.00)   &  0.14 \ (0.032)  &  0.48 \ (0.036) \\
 & & FLAP & 9.51 \ (2.58) & 0.15 \ (0.036)   & - \\
 & & FLAP$_{\sigma}$ & 3.45  \ (1.09) & 0.14 \ (0.024) & 0.40 \ (0.048)  \\
 & & BLS & 3.39 \ (0.80) & 0.13 \ (0.030) &  0.49 \ (0.053)  \\
  \hline
  0.7 & &  RVM &  3.06 \ (1.01)  & 0.18 \ (0.034)   &  0.69 \ (0.053)\\
 & & FLAP & 9.46 \ (3.47) &  0.21 \ (0.05)  & - \\
 & & FLAP$_{\sigma}$ & 2.73 \ (0.82) & 0.16 \ (0.04) & 0.56 \ (0.027)  \\
 & & BLS & 2.68 \ (0.65) & 0.16 \ (0.03) & 0.69 \ (0.042) \\
   \hline
\end{tabular}
}
\end{table}

%% file: tabel/bump.tex

\begin{table}[]
\centering
\caption{Results of the simulation study for the Bump function. For each value of SNR, 100 datasets were generated. The average number of relevance vectors, NOV, the average root mean square error, RMSE, and the average of estimated standard deviation, $\hat{\sigma}$, are reported for each method along with the corresponding sample standard deviation, SD, in the parenthesis. 
}  \vspace{0.2cm} 
\label{tab:1dbump}
\def\arraystretch{1.9}
\scalebox{0.75}{
\begin{tabular}{llllll}
   \hline
SNR & &  Method \hspace{0.1cm} & NOV (SD) \hspace{0.1cm}  & RMSE (SD) \hspace{0.1cm} & $\hat{\sigma}$ (SD) \\ 
  \hline
   10  & &  RVM & 50.2 \ (5.6)  & 0.248 \ (0.02)    & 0.256 \ (0.03)   \\
     & & FLAP & 90.7 \ (6.6)  & 0.279 \ (0.02)    & -\\
  & & FLAP$_{\sigma}$ & 43.3 \ (10.9)   &  0.257 \ (0.03)   &  0.277 \ (0.07) \\
  & & BLS & 62.5  \ (4.9) & 0.217 \ (0.02)   &    0.192 \ (0.02)       \\   
  \hline
   5  & &  RVM & 48.5 \ (5.9)  &   0.338 \ (0.03)  &  0.348 \ (0.05)   \\
     & & FLAP & 88.8 \ (4.5)  &  0.411 \ (0.03)   & -\\
  & & FLAP$_{\sigma}$ & 34.2 \ (13.9)   &  0.324 \ (0.03)   & 0.411 \ (0.09)  \\
  & & BLS & 51.2 \ (10.9) &  0.290 \ (0.03)  &     0.303 \ (0.07)         \\
  \hline
   4  & &  RVM &  36.6 \ (5.3) &   0.391 \ (0.03)  &  0.475 \ (0.06)  \\
     & & FLAP & 91.5 \ (4.9)  &  0.463 \ (0.03)   & -\\
  & & FLAP$_{\sigma}$ &  33.0 \ (15.6)  &  0.343 \ (0.04)  & 0.453 \ (0.10)  \\
  & & BLS & 37.1 \ (9.9) &  0.320 \ (0.03)  &      0.414 \ (0.08)         \\
  \hline
   3  & &  RVM & 40.8 \ (5.4)  &  0.425 \ (0.04)    &  0.487 \ (0.06)  \\
     & & FLAP &  90.3 \ (6.3) &   0.536 \ (0.03)   & -\\
  & & FLAP$_{\sigma}$ &  31.6 \ (22.2)  &  0.375 \ (0.05)   &  0.516 \ (0.14) \\
  & & BLS & 39.4 \ (13.1)  &  0.340 \ (0.04)   &    0.444 \ (0.10)         \\
  \hline 
   2  & &  RVM & 40.1 \ (5.7)  &   0.508 \ (0.05)  &  0.583 \ (0.07)  \\
     & & FLAP & 89.2 \ (5.9)  &  0.659 \ (0.04)   & -\\
  & & FLAP$_{\sigma}$ &  31.9 \ (31.2)  &  0.423 \ (0.10)  & 0.611 \ (0.21)  \\
  & & BLS & 33.9 \ (13.9) &   0.377 \ (0.06) &   0.562 \ (0.12)          \\
  \hline
 1  & &  RVM & 33.2 \ (5.3)  &   0.691 \ (0.06)   &   0.869 \ (0.10) \\
     & & FLAP & 84.6 \ (6.5)  &  0.926 \ (0.05)   & -\\
  & & FLAP$_{\sigma}$ & 23.8 \ (30.7)   & 0.515 \ (0.15)   & 0.904 \ (0.28)  \\
  & & BLS & 15.7 \ (4.2) &  0.472 \ (0.07)  &   0.967 \ (0.09)      \\    
   \hline
\end{tabular}
}
\end{table}

%% file: tabel/diabetes.tex
\begin{table}[]
\centering
\caption{Diabetes data. The estimated coefficients of the variables are listed together with the corresponding $95~\%$ Bayesian Credible Intervals, CI. }  \vspace{0.2cm}
\label{tab:diabetes}
\def\arraystretch{1.2} 
\scalebox{1}{
\begin{tabular}{llllll}
  \hline
names & Bayesian Lasso & Bayesian CI  & BLS & Bayesian CI    & Lasso \\ 
  \hline
  age & -3.080 & (-89.91,  83.01) & 0 & NA  & 0 \\ 
  sex & -203.27 & (-302.67, -103.68) & -196.87 & (-316.87, -76.87)  & -188.55 \\ 
  bmi & 523.32 & (413.92, 632.28) & 533.52 & (391.62,  675.45) & 521.18 \\ 
  map & 301.08 & (192.61,  408.54) & 304.81 & (182.91, 426.71)  & 292.36 \\ 
  tc & -144.05 & (-414.08,  66.37) & -100.60 & (-214.90, 13.70)  & -92.98 \\ 
  ldl & -18.60 & (-215.22, 195.23) & 0 & NA  & 0 \\ 
  hdl & -164.54 & (-339.62, 8.82) & -221.77 & (-373.67, -69.87)  & -220.82 \\ 
  tch & 90.31 & (-78.20,  290.18) & 0 & NA  & 0 \\ 
  ltg & 506.27 & (356.19,  663.89) & 529.17 & (373.87, 684.47)  & 508.26 \\ 
  glu & 62.18 & (-31.35, 163.78) & 20.69 & (-30.51, 71.89) & 50.20 \\ 
   \hline
\end{tabular}%
}%
\end{table}%